\pdfoutput=1
\documentclass{article} %
\usepackage{iclr2027_conference,times}

\usepackage{amsmath,amsfonts,bm}

\def\eqref#1{equation~\ref{#1}}

\def\1{\bm{1}}

\DeclareMathAlphabet{\mathsfit}{\encodingdefault}{\sfdefault}{m}{sl}
\SetMathAlphabet{\mathsfit}{bold}{\encodingdefault}{\sfdefault}{bx}{n}

\usepackage{hyperref}
\usepackage{url}
\usepackage{booktabs}
\usepackage{graphicx}
\usepackage{xcolor}
\usepackage{colortbl}
\usepackage{enumitem}
\usepackage{multirow}

\usepackage{graphicx}
\usepackage{wrapfig}        %
\usepackage{enumitem}
\usepackage{multirow}
\usepackage{makecell}
\usepackage{capt-of}      %

\usepackage[frozencache,cachedir=minted-cache]{minted}
\usepackage[most]{tcolorbox}
\definecolor{blockbg}{HTML}{EEF2EF}
\newtcolorbox{mdblock}{
    breakable,
    colback=blockbg,
    colframe=blockbg,
    arc=4pt,
    boxrule=0pt,
    left=8pt, right=8pt, top=6pt, bottom=6pt,
}

\definecolor{trajblue}{HTML}{315B7D}
\definecolor{trajbluebg}{HTML}{EDF4F8}
\definecolor{trajgreen}{HTML}{2F6B49}
\definecolor{trajgreenbg}{HTML}{EDF7F0}
\definecolor{trajred}{HTML}{9A3E3E}
\definecolor{trajredbg}{HTML}{FBEFEF}
\definecolor{trajgray}{HTML}{66727C}
\definecolor{trajgraybg}{HTML}{F3F5F6}

\makeatletter
\renewcommand{\normalsize}{%
  \@setfontsize\normalsize\@xpt\@xipt
  \abovedisplayskip      4\p@ \@plus 2\p@ \@minus 3\p@
  \abovedisplayshortskip \z@ \@plus 2\p@
  \belowdisplayskip      \abovedisplayskip
  \belowdisplayshortskip 2.5\p@ \@plus 2\p@ \@minus 2\p@
}
\makeatother
\normalsize

\makeatletter
\renewcommand{\section}{%
  \@startsection{section}{1}{\z@}%
                {-1.3ex \@plus -0.4ex \@minus -0.2ex}%
                { 0.9ex \@plus  0.2ex \@minus  0.2ex}%
                {\large\sc\raggedright}%
}
\renewcommand{\subsection}{%
  \@startsection{subsection}{2}{\z@}%
                {-1.1ex \@plus -0.4ex \@minus -0.2ex}%
                { 0.5ex \@plus  0.2ex}%
                {\normalsize\sc\raggedright}%
}
\makeatother

\setlist[itemize]{topsep=3pt, itemsep=2pt, parsep=0pt, partopsep=0pt}

\title{AMBER: Training Long-Horizon Web Agents through Append-Only Memory}

\author{\parbox{\dimexpr\textwidth-2\tabcolsep}{\centering\bfseries
Chinmay Savadikar$^{1,\dagger,*}$, Zhaoyu Zhang$^{2,\dagger}$, Mingyu Zhao$^{2}$, Shuang Xie$^{2}$, Han Li$^{2}$, \\
Tianfu Wu$^{1}$, Lingyun Wang$^{2}$ \\[4pt]
{\normalfont $^{1}$North Carolina State University \qquad $^{2}$Shopify}
}}

\usepackage[normalem]{ulem}
\newif\ifdelfinal\delfinalfalse   %

\iclrfinalcopy
\begin{document}

\maketitle
\lhead{Preprint}
\begingroup
\renewcommand{\thefootnote}{\fnsymbol{footnote}}
\footnotetext[1]{Work done as part of internship at Shopify}
\footnotetext[2]{Equal contribution}
\renewcommand{\thefootnote}{\mbox{}}\footnotetext[3]{Corresponding authors: \texttt{\{csavadi,tianfu\_wu\}@ncsu.edu}, \texttt{lingyun.wang@shopify.com}}
\endgroup

\begin{abstract}
Modern language-model agents increasingly interact with external environments over long-horizon, multi-step trajectories, where the accumulated interaction history can quickly exceed practical context budgets. To ensure reliability, agents must maintain factual information over long horizons, remember execution errors and corrective feedback, and track progress across actions. Several approaches have been proposed to achieve this without the need for maintaining the entire execution history in context, such as using the reasoning and action history, learning to maintain a fixed-size memory through an overwrite mechanism, and periodic summarization. Although overwrite memory can in principle retain anything an append-only memory can, it must learn to carry each fact through every subsequent rewrite, which is difficult to learn from sparse outcome rewards; for interactive applications like web agents, we find that trained overwrite memories delete key information required by the trajectory, as well as corrective feedback received from the environment. We introduce AMBER (\textbf{A}ppend-only \textbf{M}emory \textbf{B}ank for \textbf{E}vidence \textbf{R}etention)---a simple and scalable framework where an agent jointly learns to reason, act, and write free-form memory, while an append-only rule guarantees retention by construction. This allows AMBER to be trained end-to-end with reinforcement learning from outcome rewards without the need for extensive curated SFT data. On WebArena Lite, AMBER improves average success over overwrite-based memory by 4.09 percentage points, increases the fraction of tasks solved in five repeated runs by 4.8 percentage points, and matches an overwrite baseline trained on substantially more expensive curated supervision. AMBER achieves these improvements while maintaining a practical token budget, providing a strong balance between context efficiency, task performance, and reliable long-horizon execution. Our code is available at \url{https://github.com/agentic-foundation-modeling-research/amber}.
\end{abstract}

\section{Introduction}
\label{sec:introduction}

Language models are increasingly used as agents that interact with external
environments over long-horizon, multi-step trajectories, including web-based
deep research~\citep{openaideepresearch}, tool use~\citep{yao2025taubench}, and
browser navigation~\citep{browseruse2024}. Unlike single-turn tasks, the
information required to complete these tasks is often discovered incrementally
through interaction and exploration, and is not available upfront. An agent must therefore identify useful information from
its observations, track its execution progress, and preserve relevant evidence
until it is needed later. This raises a central question for long-horizon
agents: \emph{once an agent has identified information worth remembering, how
should that information be retained across subsequent interactions?}

The straightforward solution is to condition every decision on the complete
interaction history, including previous observations, reasoning, and
actions~\citep{yao2023react,jin2025search-r1}. However, this context grows
throughout the trajectory and quickly becomes impractical when observations
are large. Longer contexts can also make relevant information harder to
retrieve reliably~\citep{liu2024lost}. Web navigation makes this problem
particularly demanding: an agent must collect facts across pages, plan and
execute sequences of actions, track completed and pending subgoals, remember
failed attempts, recover from mistakes, and use information acquired many
steps earlier.

To make long-horizon interaction efficient, practical context management 
strategies approximate the full interaction history using some form of
compressed \textit{memory}. For reliable execution, context management mechanisms
must decide what information to \emph{write}, what previously written
information to \emph{retain}, and what information to \emph{surface}. Moreover, a useful memory mechanism should be robust to diverse
 environments and interaction settings. Several methods have been proposed to address this problem, such as retaining reasoning
and action histories while discarding past observations~\citep{webagent-r1,gandhi2025gobrowse}, learning a compact memory that is
recursively rewritten at each step~\citep{zhou2025mem1,yu2026memagent,yuan2025memsearcher}, or using external memory and summarization mechanisms~\citep{yan2025memoryr1,wu2025resum}. In this paper, we focus on the first two
mechanisms, and show that reasoning and action history alone is not sufficient, and an explicit memory is required for reliable long-horizon execution. \textit{However, when overwrite memory is trained from outcome rewards, the learned agents lose factual information and corrective feedback, and as a result repeat mistakes}.

\begin{wrapfigure}{R}{0.4\textwidth}
    \centering
    \includegraphics[width=\linewidth]{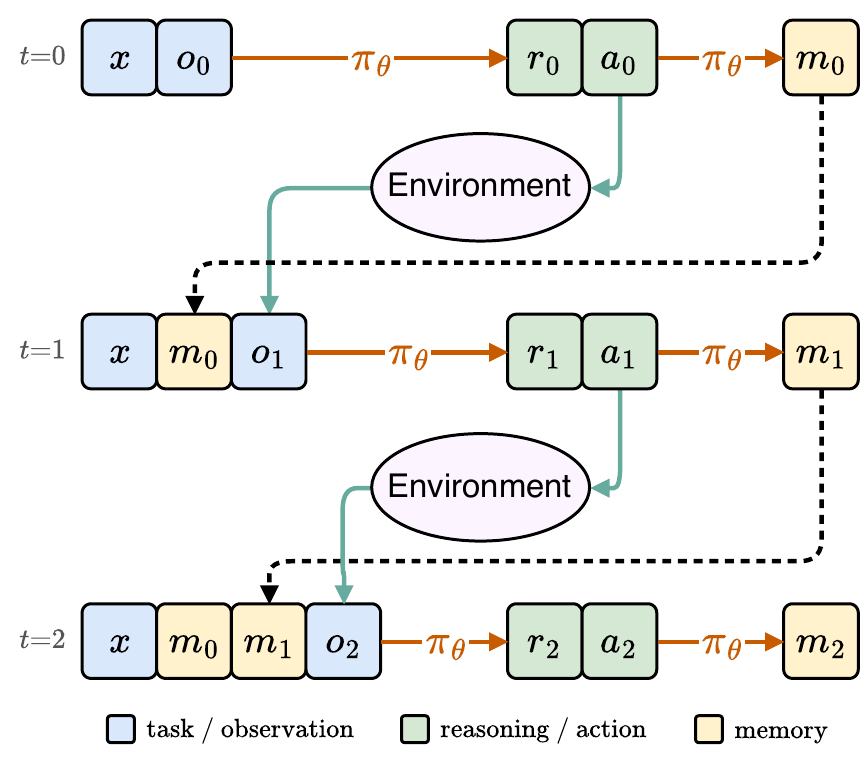}
    \caption{Overview of AMBER. At each step, the agent jointly reasons, acts, and generates a memory entry that is appended to a persistent memory bank provided as context to future steps. See text for details.}
    \label{fig:amber}
    \vspace{-6mm}
\end{wrapfigure}

We propose \textbf{AMBER} (\textbf{A}ppend-only \textbf{M}emory \textbf{B}ank
for \textbf{E}vidence \textbf{R}etention), which separates the learned decision
of \emph{what to write} from \emph{what
to retain}. At each interaction step, the agent reasons, acts, and
generates a free-form natural-language memory entry, which is appended to a
persistent memory bank (Figure \ref{fig:amber}). The policy therefore learns what information
is worth recording, while retention is determined by a simple append-only
rule. We show that AMBER can be trained jointly to optimize reasoning, action and memory generation in a scalable manner with minimal supervised data followed by end-to-end reinforcement learning from outcome rewards. To ensure scalable training, we develop and describe a rollout architecture capable of large-scale rollouts.

On WebArena~\citep{zhou2024webarena}, AMBER improves the average task success rate over overwrite memory by $4.09$ percentage
points and increases the fraction of tasks solved in five repeated runs by $4.8$ percentage points. We show that AMBER loses less information and repeats fewer execution errors. Moreover, overwrite memory matches AMBER's performance only when trained on substantially more expensive curated supervision. Together, these results suggest that the gap is one of learnability rather than capacity: the append-only prior makes durable memory easier to learn from weak outcome supervision.

Our contributions are summarized as follows, which establish AMBER as a simple, scalable method for training long-horizon web agents that balances task performance, reliability, context efficiency and supervision complexity:
\begin{itemize}[leftmargin=*]
    \item \textbf{A retention-focused view of learned agent memory.}
    We show that overwrite-style memory can result in loss of key factual information as well as corrective feedback, resulting in agents failing to solve tasks and repeating mistakes.

    \item \textbf{Append-only learned memory.}
    We introduce AMBER, which learns what information to record while using a
    deterministic append-only rule. We show that AMBER improves success rate, reliability, information retention, and shows better error correction ability than overwrite memory. While appending increases the context length in theory, we find that over trajectories in WebArena, the context length stays practical.

    \item \textbf{RL training, rollout architecture, modest supervision.} We show that AMBER can be trained in an end-to-end manner using reinforcement learning from outcome rewards. We further describe a scalable rollout infrastructure that enables this training. We show that while overwrite memory needs careful data curation for the initial SFT phase before RL, AMBER can achieve better performance without this extensive SFT, making it cheaper and easier to train.
\end{itemize}

\section{Related Work}
\label{sec:related-work}

\textbf{Training Web Agents.}
LLM-based web agents combine language understanding, planning, and interface control to complete multi-step tasks~\citep{yao2023react}, with benchmarks such as WebArena~\citep{zhou2024webarena} providing functional evaluation on realistic, reproducible websites. Two major directions have emerged for training open models in these environments. The first constructs successful trajectories for supervised fine-tuning (SFT), for example by retroactively labeling exploration~\citep{murty2024nnetnav}, replaying tutorials~\citep{xu2025agenttrek}, or framing exploration as graph search~\citep{gandhi2025gobrowse}. The second learns directly from the outcomes of online interaction, using approaches such as a self-evolving curriculum with an outcome reward model~\citep{qi2025webrl} or multi-turn reinforcement learning with binary task rewards~\citep{webagent-r1}. \citet{vattikonda2025train} study both approaches across web-agent post-training settings and find their combination important for strong performance. These systems make coupled design choices along five axes: the observation representation, the context retained across interactions, the model family, the training strategy, and the evaluation environment. Because existing systems often vary several of these axes simultaneously, their results make it difficult to isolate the effect of the context-retention strategy. Our experiments vary only the retained context while holding the other four axes fixed.

\textbf{Memory for Long-Horizon Agents.}
We use \emph{trajectory memory} to refer to information retained across steps of the same task. Several recent methods make such memory an explicit, learnable component of long-horizon reasoning, and they differ mainly in how much of the retention decision is left to the learned policy. Self-MAP augments conversational web navigation with memory and reflection~\citep{deng2024selfmap}; MEM1~\citep{zhou2025mem1} and MemSearcher~\citep{yuan2025memsearcher} jointly train reasoning and memory consolidation into a compact recurrent state; and MemAgent learns an overwrite memory for arbitrarily long segmented inputs~\citep{yu2026memagent}. In these methods, whether a fact survives is decided by the policy at every rewrite. Other systems expose learned operations over an external store. These include structured memory management trained with outcome rewards~\citep{yan2025memoryr1}, insertion and deletion as policy actions~\citep{zhang2025memact}, periodic summarization of interaction history~\citep{wu2025resum}, and learned consolidation of web-search histories~\citep{ye2025agentfold}. These operations make memory more expressive, but they also enlarge what the policy must learn, often from sparse outcome rewards alone. AMBER takes the opposite approach: the policy learns only what to write, and retention is fixed by an append-only rule.

\section{Background}
\label{sec:background}

\subsection{Web Interaction as a Partially Observable Markov Decision Process}

We model web interaction as a partially observable Markov decision process
(POMDP) $\mathcal{M}=(\mathcal{S},\mathcal{A},P,R,\Omega,O,\gamma)$. At
interaction step $t$, $s_t\in\mathcal{S}$ denotes the environment state,
$a_t\in\mathcal{A}$ is the textual action sampled by the policy,
$P(s_{t+1}\mid s_t,a_t)$ describes the website and browser dynamics,
$R_t\in\{0,1\}$ is the scalar task reward, and $\gamma$ is the discount factor.
The agent receives an observation $o_t\in\Omega$, such as a screenshot, DOM
representation, or accessibility tree, according to the observation function
$O(o_t\mid s_t,a_{t-1})$.

A straightforward approach to acting under partial observability is to
condition the policy on the complete interaction history:
\begin{equation}
    H_t=(o_0,a_0,o_1,a_1,\ldots,a_{t-1},o_t),
    \qquad
    a_t\sim\pi_\theta(\cdot\mid H_t).
    \label{eq:pomdp-history-policy}
\end{equation}
The policy is trained to maximize the expected discounted return
\begin{equation}
    J(\theta)=\mathbb{E}_{\tau\sim\pi_\theta}
    \left[\sum_{t=0}^{T}\gamma^t R_t\right].
    \label{eq:expected-return}
\end{equation}

Modern LLM-based agents implement this using the ReAct framework~\citep{yao2023react}, where an autoregressive LLM first
generates chain-of-thought reasoning $r_t$ and then an action $a_t$:
\begin{equation}
    (r_t,a_t)\sim\pi_\theta(\cdot\mid x,H_t),
    \label{eq:react-policy}
\end{equation}
where $x$ denotes the task (or goal) given to the agent, and both $r_t$ and
$a_t$ are expressed in natural language. The reasoning $r_t$ interprets the
current page, tracks intermediate goals, and motivates the next action. Unlike
$a_t$, it does not directly modify the environment, but instead conditions the
environment-affecting action that follows.

\subsection{Compressing Interaction History}
\label{sec:interaction-history}

Conditioning on the full history $H_t$ preserves all information available to
the agent at step $t$. In practice, however, retaining every past observation is
prohibitively expensive: tokenized web observations can be long, causing the
interaction history to quickly exhaust the policy model's context window.
Web agents therefore replace the full history with a compact
\textit{retained context} $C_t$ that carries information from previous steps.
For $C_t$ to serve as an effective substitute for $H_t$, the resulting policy
should approximate the behavior of the full-history policy:
\begin{equation}
    \pi_\theta(\cdot\mid x,o_t,C_t)
    \approx
    \pi_\theta(\cdot\mid x,H_t).
    \label{eq:retained-context-approximation}
\end{equation}

Concretely, at step $t$, the policy receives the current observation $o_t$
together with the retained context $C_t$ and generates the reasoning and action
\begin{equation}
    (r_t,a_t)\sim
    \pi_\theta(\cdot\mid x,o_t,C_t).
    \label{eq:retained-context-policy}
\end{equation}
An update rule $U$ then determines what information from the current interaction
is carried forward:
\begin{equation}
    C_{t+1}=U(x,C_t,o_t,r_t,a_t).
    \label{eq:retained-context-update}
\end{equation}
Thus, $U$ determines both what information persists across interactions and how
the retained context grows as the task progresses. For reliable long-horizon
execution, $C_t$ must remain a useful proxy for $H_t$. This requires the update
rule to satisfy three properties:
\begin{itemize}[leftmargin=*]
    \item \textbf{Progress tracking.} The current observation describes the
    present page, but not the agent's broader execution state: which subgoals
    have been completed, which remain unresolved, or which actions have already
    been attempted and failed. $C_t$ must preserve this information.

    \item \textbf{Information identification.} Because the retained context
    cannot include every past observation, $U$ must identify which information
    is important for completing the task and should therefore be carried
    forward.

    \item \textbf{Durability.} Once important information has been identified,
    it must remain available across subsequent interactions until it is needed
    at a future step.
\end{itemize}

\section{AMBER: Append-Only Memory}
\label{sec:method}

To satisfy the properties above, we propose AMBER (Append-only Memory Bank for
Evidence Retention), which jointly learns to reason, act, and generate step-wise
memories. AMBER instantiates the update rule
$U$ in Equation~\ref{eq:retained-context-update} using a persistent memory bank.
At each interaction step, the policy generates a free-form memory entry
containing information that may be useful later in the trajectory. The entry is
appended to the memory bank, which is provided as the retained context $C_t$ at
subsequent steps. We next describe how reasoning, action generation, and memory
generation are realized within each interaction step, and how the resulting
memory is carried forward across steps.

\subsection{Two-Turn Memory Generation and Append-Only Retention}
\label{sec:append-update}

At each environment step, AMBER interacts with the policy $\pi_\theta$ through
a two-turn conversation. At the beginning of step $t$, a new conversation is
constructed whose first user message contains the task $x$, the current
observation $o_t$, and the memory bank $C_t$. The policy first generates
reasoning $r_t$ followed by an executable action $a_t$:
\begin{equation}
    (r_t,a_t) \sim \pi_\theta(\cdot \mid x,o_t,C_t).
    \label{eq:action-policy}
\end{equation}

The conversation is then continued with a second user message instructing the
policy to generate a memory entry. Since the reasoning and action from the
first turn remain in the conversation, the same policy generates a free-form
memory $m_t$ conditioned on the preceding within-step interaction:
\begin{equation}
    m_t \sim \pi_\theta(\cdot \mid x,o_t,C_t,r_t,a_t).
    \label{eq:memory-policy}
\end{equation}
Memory generation therefore occurs after the agent has selected its action but
before it observes the outcome of that action. The entry $m_t$ can record
task-relevant information from the current observation together with the
agent's execution state and progress.

After the two model turns are complete, the environment executes $a_t$ and
returns the next observation $o_{t+1}$. The generated memory entry is appended
to the persistent memory bank:
\begin{equation}
    C_{t+1}=C_t\mathbin{\Vert}[m_t],
    \qquad\text{so that}\qquad
    C_t=[m_0,\ldots,m_{t-1}],
    \label{eq:append-memory}
\end{equation}
with $C_0$ empty. The within-step conversation is then discarded. At step
$t+1$, AMBER constructs a new two-turn conversation from $x$, $o_{t+1}$, and
the updated memory bank $C_{t+1}$. Thus, previous observations, reasoning
traces, actions, and other within-step conversation messages are not replayed
across steps; only the generated memory entries are carried forward through
$C_t$.

The update in Equation~\ref{eq:append-memory} is deterministic and contains no
learned parameters. Because each update only appends a new entry, previously
written memories cannot be modified or removed. AMBER therefore guarantees
durability by construction. The policy is responsible for deciding what
information to record and how to represent the evolving execution state,
whereas the append-only update rule determines what persists across
interactions.

Because every memory entry is retained, the size of $C_t$ grows with the
trajectory length. We show empirically that this growth remains substantially
smaller than retaining the complete interaction history and stays within
practical context limits over the horizons we evaluate.

\subsection{End-to-End Optimization Using RL}
\label{sec:grpo}
The information that should be recorded in memory can vary substantially across tasks and environments. Rather than specifying hand-crafted rules for what the
agent should remember, we learn this behavior end-to-end from task outcomes.
Specifically, we jointly optimize reasoning, action generation, and memory
generation using reinforcement learning, allowing the policy to learn what
information is useful to record and how to use the accumulated memory during
subsequent interactions. We train AMBER using outcome rewards with Group Relative Policy Optimization
(GRPO)~\citep{deepseekmath}. We observe that directly applying RL to the
pretrained instruction model produces unstable rollouts. Following
\citet{vattikonda2025train}, we therefore first perform supervised fine-tuning
(SFT) on offline trajectories to initialize the policy before RL. The resulting
SFT policy is also used as the reference policy during RL. For efficient training, we use
asynchronous RL~\citep{mnih2016asynchronous}.

\paragraph{Optimizing multi-step rollouts.}
For each training task, the behavior policy
$\pi_{\theta_{\mathrm{old}}}$ generates a group of $G$ multi-step rollouts.
Each rollout $g$ receives a single terminal reward
$R_g \in \{0,1\}$, giving
$\mathbf{R}=(R_1,\ldots,R_G)$. We compute the group-relative advantage as
\begin{equation}
    A_{g}
    = \frac{R_{g}-\operatorname{mean}(\mathbf{R})}
           {\operatorname{std}(\mathbf{R})+\delta},
    \label{eq:group-relative-advantage}
\end{equation}
where $\delta$ is a small constant for numerical stability. Every
interaction step in rollout $g$ is assigned the same advantage $A_{t,g} = A_g$, and treated as an independent training sample during policy-gradient optimization. For the two-turn interaction defined in
Section~\ref{sec:append-update}, a step contains all assistant-generated tokens from
the reasoning $r_{t,g}$, action $a_{t,g}$, and memory $m_{t,g}$. Template and
user-message tokens inserted between the two assistant turns are masked from
the loss. Consequently, every assistant-generated token at step $t$ receives
the same rollout-level advantage $A_g$.

Let $\mathcal{Y}_{t,g}$ denote the set of assistant-generated tokens at step
$t$ of rollout $g$, including reasoning, action, and memory tokens. For token
$i \in \mathcal{Y}_{t,g}$, we define the importance-sampling ratio
\begin{equation}
    \eta_{t,g}^{(i)}
    =
    \frac{
        \pi_{\theta}
        \left(
            y_{t,g}^{(i)}
            \mid y_{t,g}^{(<i)},\, x,\, o_{t,g},\, C_{t,g}
        \right)
    }{
        \pi_{\theta_{\mathrm{old}}}
        \left(
            y_{t,g}^{(i)}
            \mid y_{t,g}^{(<i)},\, x,\, o_{t,g},\, C_{t,g}
        \right)
    }.
\end{equation}
The resulting GRPO objective is
\begin{align}
\mathcal{J}(\theta)
&=
\mathbb{E}_{x \sim \mathcal{D}(x),\tau \sim \pi_{\theta_{\mathrm{old}}}}
\Bigg[
\frac{1}{G}
\sum_{g=1}^{G}
\frac{1}{T_g}
\sum_{t=0}^{T_g-1}
\frac{1}{|\mathcal{Y}_{t,g}|}
\sum_{i \in \mathcal{Y}_{t,g}}
\min\Big(
    \eta_{t,g}^{(i)} A_g,
    \operatorname{clip}
    \big(
        \eta_{t,g}^{(i)},
        1-\epsilon_l,
        1+\epsilon_u
    \big) A_g
\Big)
\Bigg] \notag \\
&-
\beta\,
\mathbb{KL}
\left(
    \pi_\theta
    \,\|\, 
    \pi_{\mathrm{ref}}
\right),
\label{eq:amber-grpo}
\end{align}
where $\epsilon_l$ and $\epsilon_u$ are the lower and upper clipping ranges,
respectively, and $\beta$ controls the KL penalty to the SFT reference policy.

Thus, AMBER receives no separate memory-specific reward during RL. The memory
writer is optimized jointly with reasoning and action generation through the
terminal outcome reward, allowing useful memory-writing behavior to emerge from
task performance rather than from hand-crafted memory rules.

\subsection{Rollout Infrastructure}
\label{sec:rollout-infrastructure}

\begin{figure}[t]
    \centering
    \includegraphics[width=0.8\linewidth]{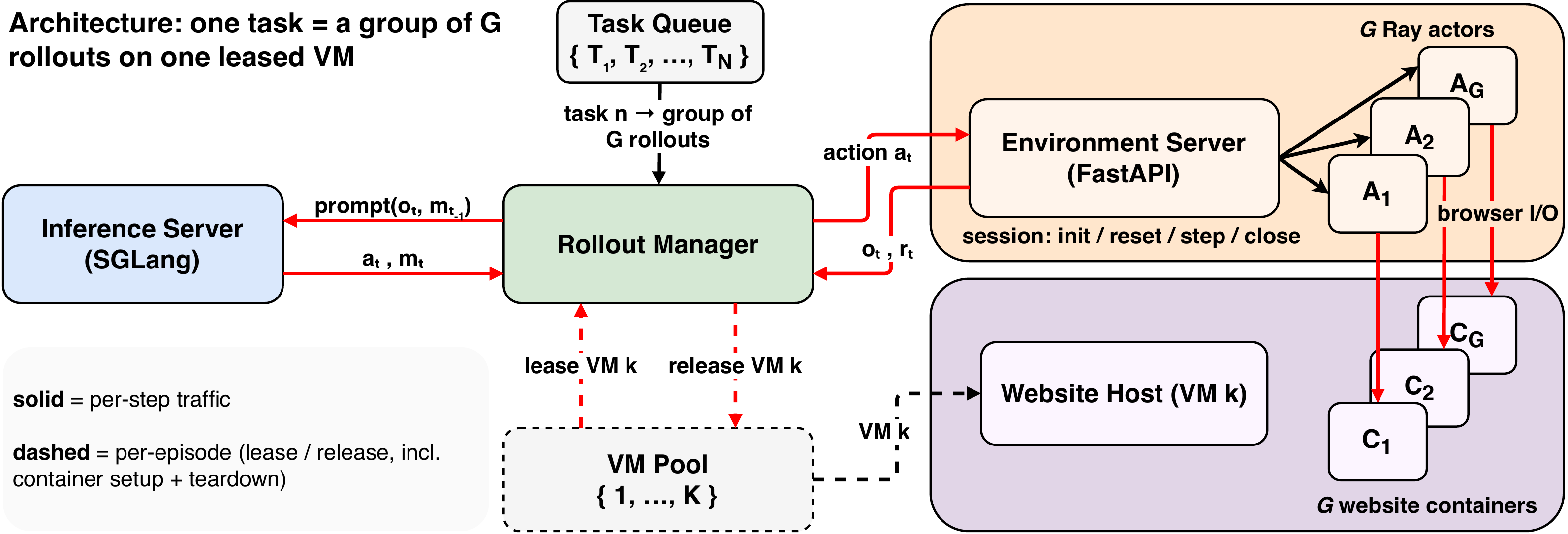}
    \caption{Rollout architecture. Each task is expanded into a group of $G$ rollouts on a leased website VM, and each rollout gets its own Ray-managed browser session and website container. Red arrows denote HTTP connections. Solid arrows show per-step traffic between the inference server, rollout manager, and environment server, and dashed arrows show per-episode VM lease and release.}
    \label{fig:rollout-architecture}
\end{figure}

Efficient rollout generation for web-agent RL requires handling several sources
of isolation and latency. First, each rollout needs an independent browser session 
so that browser state, such as cookies, does not leak across rollouts. Second, some 
WebArena tasks modify persistent website state, requiring separate website containers 
across concurrent rollouts. Third, browser interaction is often substantially slower 
than model inference, so synchronizing all rollouts at every step can lead to poor 
GPU utilization. We therefore separate model inference, browser execution, and website 
serving into independently scalable components. Figure~\ref{fig:rollout-architecture} 
shows the resulting architecture. Each rollout, corresponding to a multi-step episode, 
is assigned an isolated browser session and website container.

\paragraph{Remote browser execution and rollout isolation.}
We use Ray~\citep{ray} to manage concurrent browser sessions and reward models. The 
Ray actors are exposed through an environment server implemented in FastAPI. At each 
environment step $t$, the rollout manager sends the action $a_t$ to the environment server 
over HTTP. The server routes the request to the Ray actor associated with that rollout, 
which executes the action in the browser and computes the reward $R_t$. The resulting 
observation $o_{t+1}$ is then returned to the rollout manager through the environment 
server. Each rollout is assigned a unique identifier, which is used to route requests 
to the corresponding Ray actor. For tasks that modify website state, the rollout is also 
assigned a dedicated website container hosted on a separate virtual machine, preventing 
changes made by one rollout from affecting another. In practice, each Ray actor creates 
a BrowserGym~\citep{browsergym} environment to manage the browser interaction and rewards.

\paragraph{Asynchronous environment interaction.}
Rollouts execute asynchronously at the interaction-step level. In particular, rollouts do 
not wait for one another after each environment step. While one rollout is waiting for browser 
execution, page loading, or reward computation, another rollout whose environment step has 
completed can issue a model-inference request for its next action. Since both the inference 
server, implemented with SGLang, and the environment server are asynchronous HTTP servers, 
environment interaction and model inference can overlap across different rollouts. This 
reduces idle inference time caused by slow browser execution and allows the rollouts to 
progress independently. The complete implementation details are provided in
Appendix~\ref{sec:rollout-infrastructure-details}.

\section{Experiments}
\label{sec:experiments}

\textbf{Environment.}
We use WebArena~\citep{zhou2024webarena} as the training and evaluation
environment. WebArena provides five self-hosted websites: Shopping, Shopping
Admin (a content management system), Reddit, GitLab, and Maps, together with
$812$ curated tasks verified through functional evaluation. We evaluate on
WebArena Lite~\citep{liu2024visualagentbench}, a curated subset of these tasks,
and use the remaining tasks during reinforcement learning. Early in our
experiments, we found that GitLab produced unstable container setups when many
instances were launched in parallel, and therefore exclude it from our experiments.

\textbf{Input Representation and Environment Interaction.}
We use the flattened accessibility tree restricted to elements in the current viewport, together with open-tab metadata. We use BrowserGym~\citep{browsergym} for
accessibility-tree extraction and browser interaction. We cap RL rollouts at 20 interaction steps and evaluation at 30 steps.

\textbf{Curating synthetic data for Supervised Fine-Tuning.} For the SFT stage, we use trajectories released by Go-Browse~\citep{gandhi2025gobrowse},
which contain accessibility-tree observations coupled with step-wise reasoning and action traces for supervision. We
construct step-wise memory labels required by AMBER by retrospectively annotating successful
trajectories using Pi coding
agent~\citep{piCodingAgent} with GPT 5.5. We provide the full details of the data generation along with the prompts in
Appendix~\ref{sec:synthetic-data-generation}.%

\textbf{Evaluation.} We evaluate AMBER across five independent runs with different seeds using three
metrics: \textbf{(1) Success rate.} The success rate for each website, averaged over tasks and the five runs (Section~\ref{sec:memory-results}). \textbf{(2) Repeated-run reliability.} For each task, we count the number of successful runs and report the fraction of tasks solved in at least $k$ of five runs, for $k=1,\ldots,5$ (Section~\ref{sec:append-robustness}). \textbf{(3) Context length.} The token length of the action prompt, including the retained context $C_t$ (Section~\ref{sec:context-cost}).

\textbf{Baselines.} We compare AMBER against two established context-management strategies: \textit{Reasoning+Action History}, which retains past reasoning and actions (e.g., WebAgent-R1~\citep{webagent-r1}, Go-Browse~\citep{gandhi2025gobrowse}), and \textit{Overwrite Memory}, which maintains a learned memory through overwrite updates (e.g., MemAgent~\citep{yu2026memagent}, MemSearcher~\citep{yuan2025memsearcher}). We use WebAgent-R1 and MemAgent as representative methods from these respective categories, re-implementing both baselines with reference to their official implementations. All methods use the Qwen 3.5~\citep{qwen35blog} model family under a common experimental setup. Experimental details and hyperparameters are provided in Appendix~\ref{sec:rl-hyperparameters}.

\subsection{Success Rate}
\label{sec:memory-results}

Table~\ref{tab:success-rates} reports success rates on WebArena Lite, averaged
over five independent runs. We make the following observations:

\begin{itemize}[leftmargin=*]
    \item \textbf{Zero-shot evaluation.}
    Under a zero-shot setting, Qwen 3.5 9B achieves a 34.22\% success rate under the full-context protocol, which drops to 21.82\% under the AMBER protocol. The same trend is observed for 27B. This shows that Qwen 3.5 is capable of reasoning and acting in web environments, but cannot generate compressed step-wise memory reliably.

    \item \textbf{Benefit of explicit memory.}
    MemAgent, which maintains an explicit learned memory through overwrite
    updates, improves over the reasoning-and-action history of WebAgent-R1 by
    $4.41$ percentage points. This suggests that explicitly training the agent
    to record task-relevant information is beneficial for long-horizon
    interaction, rather than relying on reasoning generated primarily to support
    the next action.

    \item \textbf{Append-only retention improves success rate.}
    AMBER achieves the highest average success rate, outperforming MemAgent by $4.09$ percentage points with Qwen 3.5 9B, and by 2.05 points with 27B. AMBER with Qwen 3.5 27B outperforms Haiku 4.5 by 2.1 points.
\end{itemize}

\begin{table}[b]
\centering
\caption{Success rate (\%) on WebArena Lite, averaged over five evaluation runs.}
\label{tab:success-rates}
\small
\setlength{\tabcolsep}{3.5pt}
\resizebox{0.7\linewidth}{!}{
\begin{tabular}{lrrrrrr}
\toprule
\textbf{Model} & \textbf{Shopping} & \textbf{Admin} & \textbf{Reddit} & \textbf{Map} & \textbf{Multi-site} & \textbf{Average} \\
\midrule
\multicolumn{7}{c}{\cellcolor[HTML]{F0F0FF}\textit{Closed Models/Zero-Shot with AMBER}} \\
GPT-5-mini & 28.68 & 38.10 & 29.41 & 25.64 & 27.78 & 30.71 $\pm$ 1.11 \\
Haiku 4.5 & 44.96 & 62.86 & 52.94 & 25.64 & 38.89 & 46.72 $\pm$ 3.24 \\
Qwen 3.5 9B & 27.05 & 28.00 & 10.72  & 14.10 & 16.67 & 21.82 $\pm$ 2.97 \\
Qwen 3.5 27B & 37.21 & 43.43 & 42.35  & 17.69 & 40.00 & 35.75 $\pm$ 4.19 \\
\midrule
\multicolumn{7}{c}{\cellcolor[HTML]{F0F0FF}\textit{Qwen 3.5 9B}} \\
Full Context (Zero-Shot)   & 36.08 & 40.20 & 48.24 & 14.00 & 33.33 & 34.22 $\pm$ 1.30 \\
WebAgent-R1 & 29.46 & 34.29 & 39.22 & 10.26 & 33.33 & 28.35 $\pm$ 0.64 \\
MemAgent  & 28.84 & 41.71 & 44.71 & 17.69 & 40.00 & 32.76 $\pm$ 1.54 \\
AMBER (ours)     & 29.30 & 43.43 & 61.18 & 23.08 & 43.33 & \textbf{36.85} $\pm$ 1.68 \\
\midrule
\multicolumn{7}{c}{\cellcolor[HTML]{F0F0FF}\textit{Qwen 3.5 27B}} \\
Full Context (Zero-Shot)   & 47.91 & 58.29 & 55.29 & 21.54 & 26.67 & 45.35 $\pm$ 1.73 \\
WebAgent-R1 & 43.26& 54.29 & 76.47 & 14.62 & 36.67 & 44.57 $\pm$ 3.46 \\
MemAgent  & 48.84 & 56.57 & 71.76 & 15.38 & 40.00 & 46.77 $\pm$ 3.18 \\
AMBER (ours)     & 49.77 & 56.57 & 70.59 & 23.85 & 43.33 & \textbf{48.82} $\pm$ 1.36 \\
\bottomrule
\end{tabular}}
\end{table}

\subsection{Reliability Across Repeated Runs}
\label{sec:append-robustness}

\begin{table}[t]
\begin{minipage}[t]{0.5\linewidth}
\vspace{0pt}%
\centering
\caption{Repeated-run reliability across five evaluation runs. Each column reports the percentage of tasks solved in at least the indicated number of runs (e.g., $\geq 3/5$ denotes tasks solved in at least three of five runs).}
\label{tab:reliability-curve}
\small
\setlength{\tabcolsep}{3.5pt}
\resizebox{\linewidth}{!}{%
\begin{tabular}{lrrrrr}
\toprule
 & $\geq 1/5$ & $\geq 2/5$ & $\geq 3/5$ & $\geq 4/5$ & $5/5$ \\
\midrule
\multicolumn{6}{c}{\cellcolor[HTML]{F0F0FF}\textit{Qwen 3.5 9B}} \\
WebAgent-R1 & 43.3\% & 34.6\% & 25.2\% & 24.4\% & 15.7\% \\
MemAgent & 48.8\% & 39.4\% & 34.6\% & 25.2\% & 15.7\% \\
AMBER & \textbf{53.5}\% & \textbf{41.7}\% & \textbf{36.2}\% & \textbf{32.3}\% & \textbf{20.5}\% \\
\midrule
\multicolumn{6}{c}{\cellcolor[HTML]{F0F0FF}\textit{Qwen 3.5 27B}} \\
WebAgent-R1 & \textbf{62.2}\% & 49.6\% & 44.1\% & 36.2\% & 30.7\% \\
MemAgent & 59.1\% & 52.0\% & 48.0\% & \textbf{42.5}\% & 32.3\% \\
AMBER & 61.4\% & \textbf{55.1}\% & \textbf{50.4}\% & 41.7\% & \textbf{35.4}\% \\
\bottomrule
\end{tabular}}
\end{minipage}\hfill
\begin{minipage}[t]{0.45\linewidth}
\vspace{0pt}
\centering
\includegraphics[width=0.93\linewidth]{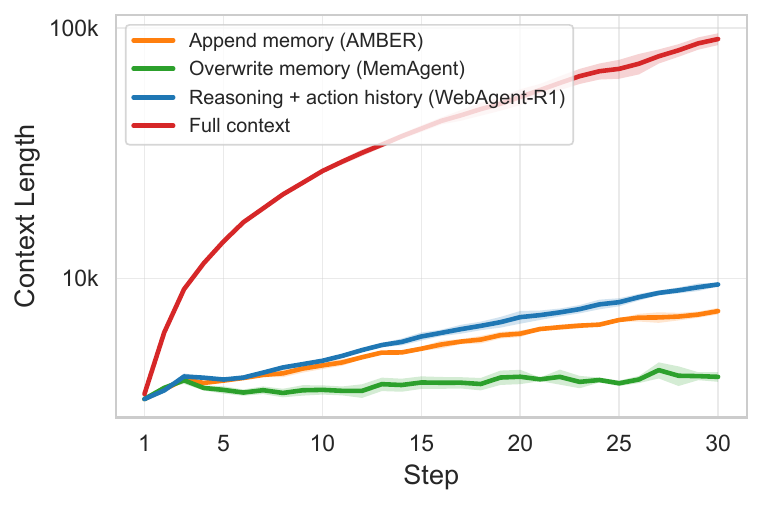}
\captionof{figure}{Mean action-prompt length by interaction step, averaged over five runs. The context length is plotted in log scale.}
\label{fig:context-lengths}
\end{minipage}
\end{table}
Table~\ref{tab:reliability-curve} evaluates how consistently each method solves
the same task across the five runs. Each column applies a progressively stricter
criterion, from solving a task at least once ($\geq 1/5$) to solving it in all
five runs ($5/5$).
For the 9B model, AMBER leads at all five reliability thresholds. For the 27B
model, AMBER leads at $\geq 2/5$, $\geq 3/5$, and $5/5$, while WebAgent-R1 leads
at $\geq 1/5$ and MemAgent at $\geq 4/5$.

\textbf{Tasks brought within reach.}
For the 9B model, both memory-based methods increase the fraction of tasks solved at least once, from 43.3\% with reasoning history to 48.8\% with overwrite memory and 53.5\% with AMBER.

\textbf{Tasks solved repeatedly.}
At the majority-run threshold ($\geq 3/5$), success increases from $25.2\%$
with reasoning history to $34.6\%$ with overwrite memory and $36.2\%$ with
AMBER for the 9B model. At the strictest $5/5$ threshold, AMBER reaches $20.5\%$ with 9B and $35.4\%$ with 27B, outperforming both baselines. \textit{Thus, AMBER's main advantage appears in
solving tasks consistently across repeated runs rather than simply solving
more tasks once.}

\subsection{Context Length}
\label{sec:context-cost}

Since AMBER preserves every memory entry, its context length grows with the
trajectory length, whereas overwrite memory remains bounded.
Figure~\ref{fig:context-lengths} quantifies this difference using the fully
rendered action prompt. At step $30$, AMBER reaches $7.4$k tokens compared
with $4$k for MemAgent and $9$k for WebAgent-R1. However, the context length for AMBER stays well below the full-context baseline, which reaches $90$k tokens by step 30, and within the model's maximum context length. Although AMBER results in a larger context length, it remains practical. \textit{AMBER trades a moderate increase in context length for stronger and more reliable task execution}.

\subsection{Supervision Strength vs.\ Retention Prior}
\label{sec:refined-overwrite}

One possible explanation for the weaker performance of overwrite memory is that
RL cannot learn to write a cumulative and lossless memory without stronger supervised data. To test whether better supervised data during the initial SFT phase can reduce the gap, we refine the supervision targets for the generated memory using a pipeline consisting of a judge model that penalizes lossy compression and ensures that the memory is cumulative
(see Appendix~\ref{sec:refined-overwrite-data} for details). As shown in Table~\ref{tab:success-rates-refined}, refined supervision improves
MemAgent's average success rate from $32.76\%$ to $35.59\%$. The same pattern appears in repeated-run
reliability (Table~\ref{tab:reliability-curve-refined}): AMBER matches the
refined overwrite model at three of the five thresholds, while refined
overwrite leads at the remaining two. We use Qwen 3.5 9B for the experiments.

These results show that stronger SFT supervision can improve
overwrite memory. However, obtaining
this improvement requires a more complex and expensive data-generation pipeline. AMBER
achieves this level of performance without requiring this additional refinement procedure.

\begin{table}[!h]
\centering
\small
\setlength{\tabcolsep}{3.5pt}
\begin{minipage}[t]{0.54\linewidth}
\centering
\caption{Effect of refined training data on success rate, compared with the original MemAgent and AMBER.}
\label{tab:success-rates-refined}
\resizebox{\linewidth}{!}{%
\begin{tabular}{lrrr}
\toprule
& \textbf{MemAgent} & \textbf{+ Refined data} & \textbf{AMBER (ours)} \\
\midrule
Shopping    & 28.84 & 27.44 & 29.30 \\
Admin       & 41.71 & 50.86 & 43.43 \\
Reddit      & 44.71 & 56.47 & 61.18 \\
Map         & 17.69 & 14.62 & 23.08 \\
Multi-site  & 40.00 & 36.67 & 43.33 \\
\midrule
Average & 32.76 $\pm$ 1.54 & 35.59 $\pm$ 1.75 & \textbf{36.85} $\pm$ 1.68 \\
\bottomrule
\end{tabular}}
\end{minipage}\hfill
\begin{minipage}[t]{0.43\linewidth}
\centering
\caption{Effect of refined training data on repeated-run reliability, compared with the original MemAgent and AMBER.}
\label{tab:reliability-curve-refined}
\resizebox{\linewidth}{!}{%
\begin{tabular}{lrrr}
\toprule
& \textbf{MemAgent} & \textbf{+ Refined data} & \textbf{AMBER} \\
\midrule
$\geq 1/5$ & 48.8\% & \textbf{54.3}\% & 53.5\% \\
$\geq 2/5$ & 39.4\% & 39.4\% & \textbf{41.7}\% \\
$\geq 3/5$ & 34.6\% & 34.6\% & \textbf{36.2}\% \\
$\geq 4/5$ & 25.2\% & 28.3\% & \textbf{32.3}\% \\
$5/5$      & 15.7\% & \textbf{21.3}\% & 20.5\% \\
\bottomrule
\end{tabular}}
\end{minipage}
\end{table}

\newpage
\section{Trajectory Analysis}
\label{sec:append-analysis}
\label{sec:recursive-compression-loss}

\begin{wraptable}{r}{0.35\textwidth}
\vspace{-4mm}
\centering
\caption{Percentage of trajectories retaining environment feedback in memory after $k$ steps.}
\label{tab:memagent-lesson-survival}
\small
\setlength{\tabcolsep}{3.5pt}
\resizebox{\linewidth}{!}{%
\begin{tabular}{l|ccccc}
\toprule
$k$ & \textbf{1} & \textbf{2} & \textbf{3} & \textbf{4} & \textbf{5} \\
\midrule
\% trajectories & 100.0 & 25.6 & 7.5 & 4.5 & 0.8 \\
\bottomrule
\end{tabular}%
}
\vspace{-2mm}
\end{wraptable}

\begin{figure}[t]
    \centering
    \includegraphics[width=0.8\linewidth]{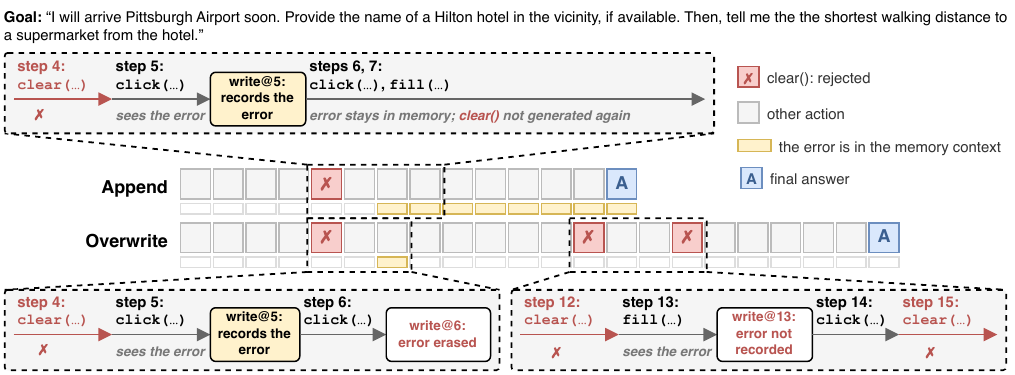}
    \caption{Error recovery on a Maps task. The environment supports \texttt{fill} but not \texttt{clear}. Both agents issue \texttt{clear}, see the error, and record it in memory. MemAgent's next overwrite erases the record, and it issues \texttt{clear} repeatedly. AMBER's appended entry persists, and it never repeats the action. See Figure~\ref{fig:error-recovery-task-33} for the detailed trajectory.}
    \label{fig:error-recovery}
\end{figure}

In this section, we analyze the trajectories of agents with overwrite vs. append memories for a more nuanced understanding of the failures. We observe two main patterns which highlight the failure modes of overwrite memory and how append memory mitigates them:

\textbf{Recovery after execution errors.} Overwrite memory can erase error feedback from the environment, potentially causing the agent to repeat previously rejected actions. We focus on invalid actions, for which the environment provides clear, actionable feedback that can guide recovery. Table~\ref{tab:memagent-lesson-survival} shows that this feedback is lost from memory after four steps, and Figure~\ref{fig:error-recovery} illustrates the repetition of previously rejected actions. To quantify this behavior, we define an \emph{event} as the first rejection of an invalid action within an episode and count how often the agent repeats that action afterward. Averaged over five seeds, AMBER re-issues an invalid action 1.29 times per event, compared with 1.95 for MemAgent. Both methods heed most corrections, but MemAgent re-issues the invalid action six or more times in 15.4\% of events, compared with 3.3\% for AMBER. While this analysis focuses on invalid actions, \textit{it shows that append memory can retain and act on feedback better than overwrite memory}.

\textbf{Information loss.} Overwrite memory can also drop evidence that was already recorded correctly, making it unavailable when needed later. We call this failure mode \emph{recursive-compression loss} (Figure~\ref{fig:task23-trajectories}; more examples in Appendix~\ref{app:additional-trajectories}).

\begin{figure}[t]
    \centering
    \includegraphics[width=0.8\linewidth]{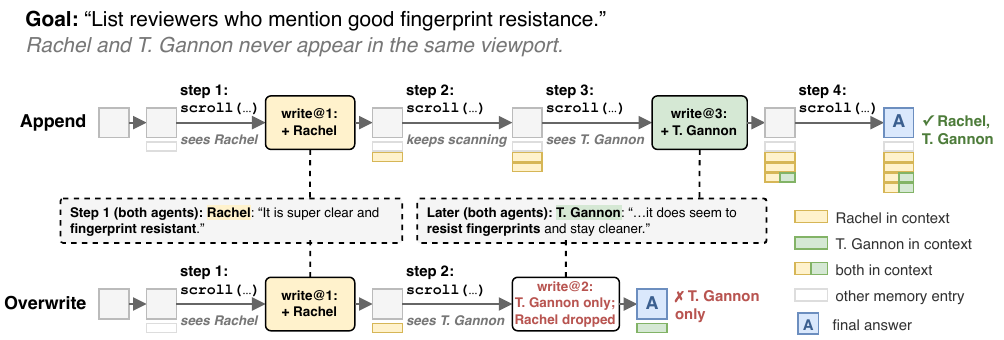}
    \caption{Recursive-compression loss. The task asks for reviewers who mention good fingerprint resistance, and the two matching reviews never share a viewport. Both agents record Rachel after the first scroll. The overwrite agent's rewrite after it sees T.\ Gannon drops Rachel, and it answers with T.\ Gannon only. AMBER's original entry persists alongside the new one, and it answers correctly with both reviewers.}
    \label{fig:task23-trajectories}
\end{figure}

\vspace{-2mm}
\section{Conclusion}
\label{sec:conclusion}
\vspace{-2mm}
In this paper, we study context retention in long-horizon web agents and propose AMBER (Append-only Memory Bank for Evidence Retention). AMBER separates the learned decision of what to record from the deterministic decision of what to retain: the policy writes a free-form memory entry at each step, and an append-only rule preserves every entry. We show that AMBER can be trained end-to-end with reinforcement learning from outcome rewards after minimal supervised fine-tuning. On WebArena Lite, AMBER improves average success over overwrite memory across 9B and 27B model sizes, and achieves stronger repeated-run reliability. Our trajectory analysis shows that overwrite memory can lose recorded evidence and error feedback. While refined supervision narrows the gap for overwrite memory, it requires an expensive data curation pipeline, and AMBER matches it without this additional curation. Thus, AMBER is a simple, scalable method for training long-horizon web agents that balances task performance, reliability, context efficiency and supervision complexity.

\section{AI Disclosure}
In this work, we use generative AI tools for the following:
\begin{itemize}
    \item Writing code, which is verified by the lead author.
    \item Drafting parts of the manuscript, which is proofread and rephrased by the authors.
    \item Synthetic data generation (details are provided in the relevant sections)
    \item Data analysis and trajectory analysis.
\end{itemize}

We take responsibility for the final content of this work, including text, claims or artifacts produced with the aid of generative AI.

\bibliography{iclr2027_conference}
\bibliographystyle{iclr2027_conference}

\appendix

\section{Limitations}
\label{sec:limitations}

AMBER
retains every generated memory entry, so its context grows linearly with the
number of interaction steps. Over the horizons we evaluate, this growth remains
well within the model's context window and is comparable to retaining the
reasoning history. At substantially longer horizons, however, some form of
selection or consolidation may become necessary. How to introduce such
operations while preserving information that may become relevant much later in
a trajectory remains an open question. 

In both append and overwrite memory, we observe that models sometimes hallucinate when generating the memory. This could potentially be an effect of the model being trained with reinforcement learning solely using outcome rewards. We leave addressing this to future work.

Finally, our information loss analysis is qualitative. The examples in
Section~\ref{sec:append-analysis} demonstrate that information can be recorded
correctly and subsequently lost through memory rewriting, but they are
examples rather than an estimate of how frequently this failure mode occurs.
A systematic characterization of memory failures across tasks and trajectories
would provide a more complete picture of when different retention strategies
succeed or fail.

\section{Broader Impact}
\label{sec:broader-impact}
AMBER improves evidential durability and auditability, but append-only retention should not imply indefinite storage in deployed systems. Persistent entries may contain sensitive, stale, incorrect, or adversarial information. Practical systems should separate the immutable trajectory representation studied here from policies governing storage duration, prompt-time selection, user inspection and deletion, and redaction of private information. Web agents also require safeguards against persistent prompt injection and consequential unauthorized actions. Our experiments use self-hosted benchmark environments and involve no real users.

\section{Extended Related Work and Design Space}
\label{sec:design-space}

This appendix expands Section~\ref{sec:related-work}.

\paragraph{Trajectory supervision and outcome-based adaptation.}
Trajectory-based methods obtain demonstrations without relying entirely on
manual annotation.  NNetNav explores websites and retroactively labels action
sequences~\citep{murty2024nnetnav}, AgentTrek converts tutorials into task
specifications and replays them in real environments~\citep{xu2025agenttrek},
and Go-Browse frames exploration as graph search to collect successful
long-horizon trajectories for SFT~\citep{gandhi2025gobrowse}.  Outcome-based
methods instead optimize the policy during interaction.  WebRL introduces a
self-evolving curriculum and an outcome reward model~\citep{qi2025webrl}, and
WebAgent-R1 performs multi-turn RL with binary task rewards after a
behavior-cloning warm start~\citep{webagent-r1}.  More broadly,
\citet{vattikonda2025train} study SFT and on-policy RL across web-agent
post-training settings and find their combination important for strong
performance.

\paragraph{The design space.}
These works make coupled choices along five axes: the observation representation
of the current environment, the context retained across interactions, the model
family, the training strategy, and the environment in which the system is
evaluated.  The evaluation environment deserves particular attention, because it
determines how much pressure a retention rule is actually under.
Table~\ref{tab:prior-design-space} organizes representative systems along these
axes as a conceptual map, not a performance comparison; the systems differ in
model, action space, and training budget.  Our experiments vary only the second
axis.  We hold the observation representation, model family, training strategy,
and evaluation environment fixed, and compare retained-context designs in one
shared setting.

\begin{table*}[h]
\centering
\caption{Representative combinations of evaluation environment, observation
input, retained context, and training.  Recurrent overwrite memory has so far
been validated on single-template or non-interactive settings, and not on
diverse web navigation.  The table organizes design choices; it does not imply
that results are directly comparable across different tasks and environments.}
\label{tab:prior-design-space}
\small
\setlength{\tabcolsep}{4pt}
\resizebox{\linewidth}{!}{
\begin{tabular}{llllll}
\toprule
System & Evaluation environment & Observation input & Retained info. & Retention & Training \\
\midrule
Go-Browse & WebArena & Accessibility tree & Reasoning history & Accumulate & SFT \\
WebAgent-R1 & WebArena & Simplified HTML & Reasoning history & Accumulate & SFT + RL \\
MEM1 & WebShop, multi-hop QA & Custom text & Explicit memory & Recurrent overwrite & RL \\
MemAgent & Long-document QA & Text segments & Explicit memory & Recurrent overwrite & RL \\
AMBER (ours) & WebArena & Accessibility tree & Explicit memory & Append-only & SFT + RL \\
\bottomrule
\end{tabular}}
\end{table*}

\paragraph{Experiential memory.}
Trajectory memory, the subject of this paper, differs from \emph{experiential
memory}, which transfers knowledge across tasks or sessions.  Reflexion stores
verbal feedback from failed attempts~\citep{shinn2023reflexion}; ExpeL extracts
reusable natural-language insights~\citep{zhao2024expel}; Voyager accumulates
executable skills~\citep{wang2023voyager}; and MemGPT manages multiple memory
tiers beyond the model context~\citep{packer2023memgpt}.  These systems address
which past experience should be retained or retrieved for future problems.  We
instead study whether evidence generated during one ongoing task remains
available until that task terminates.

\section{Rollout Infrastructure Details}
\label{sec:rollout-infrastructure-details}

Our rollout infrastructure separates policy inference, browser execution, and
website serving so that each component can be provisioned independently.

\paragraph{Remote browser execution.}
Browser sessions run on dedicated environment servers and are accessed through
an HTTP interface for session creation, reset, interaction, and termination.
This transport layer is hidden behind a thin Gym-style client, so the rollout
manager in the trainer interacts with each remote browser through the same
environment interface as a local environment and never issues HTTP requests
directly. Before being returned to the policy, each observation is reduced to
the flattened accessibility tree, open-tab titles and URLs, and the most recent
action error. This abstraction separates rollout logic from browser placement,
allowing the same rollout code to operate against any reachable environment
server.

\paragraph{Step-level asynchronous execution.}
Each rollout progresses independently through alternating policy-inference and
environment-interaction steps. Both operations are issued asynchronously, so
rollouts need not synchronize after each interaction. For example, while one
rollout waits for a browser action, page load, or task validation to complete,
another rollout whose previous environment step has finished can perform model
inference for its next action. This interleaving allows environment latency and
policy inference to overlap across concurrent rollouts.

\paragraph{Browser isolation.}
Each rollout is assigned an independent BrowserGym \citep{browsergym} session hosted by a
Ray~\citep{ray} actor. Concurrent rollouts therefore do not share browser state,
cookies, or navigation history. The environment server tracks active sessions,
limits the number of concurrent browsers, and reclaims sessions after
termination or prolonged inactivity. Browser failures are contained within the
corresponding rollout rather than affecting other active sessions.

\paragraph{Website hosting and state isolation.}
Website services run on hosts separate from browser execution and policy
inference. This separation allows website capacity to scale independently and
is particularly important for tasks that modify persistent website state. Such
rollouts are assigned isolated copies of the required mutable services so that
writes made by one rollout cannot affect another.

For WebArena, we share services whose state does not need to be isolated and
replicate only the mutable services required by a task. Each website host
maintains a set of rollout slots, where a slot contains dedicated containers
for the mutable services assigned to that rollout. A slot is returned to the
available pool only after these containers have been torn down, preventing
state from leaking between successive rollouts. Because no WebArena task
modifies more than two mutable services, our configuration supports eight
concurrent isolated rollout slots per website host.

\begin{figure}
    \centering
    \includegraphics[width=0.9\linewidth]{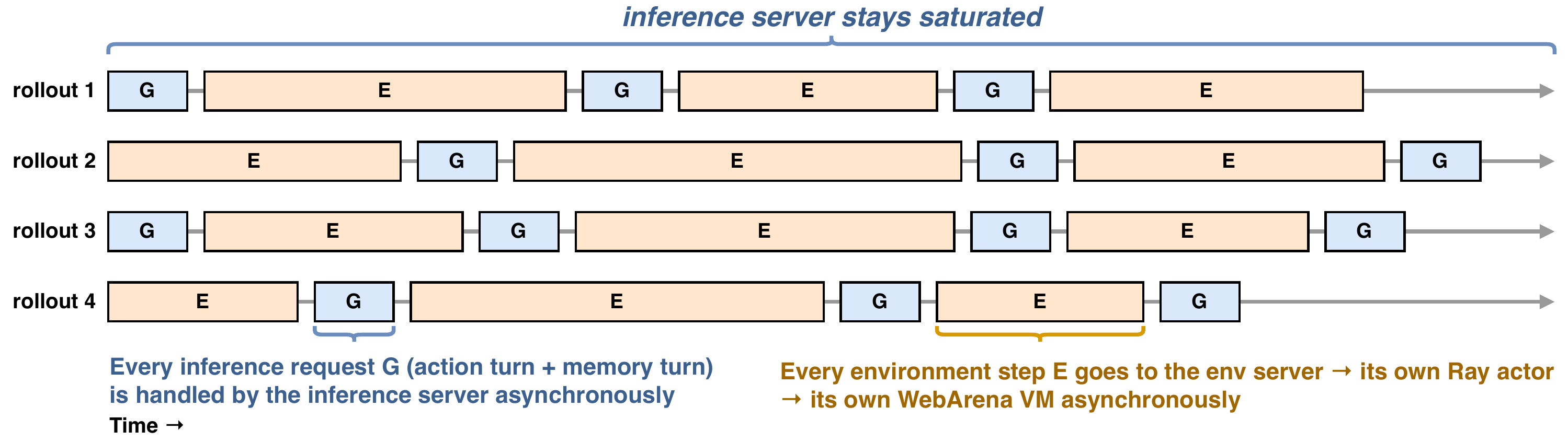}
    \caption{Asynchronous Rollouts: Each rollout is a multi-step episode with each step accessing the environment and inference asynchronously.}
    \label{fig:async-requests}
\end{figure}

\clearpage
\section{Recovering from Execution Errors}
\label{app:error-recovery}

\begin{figure}[!h]
    \centering
    \includegraphics[width=0.9\linewidth]{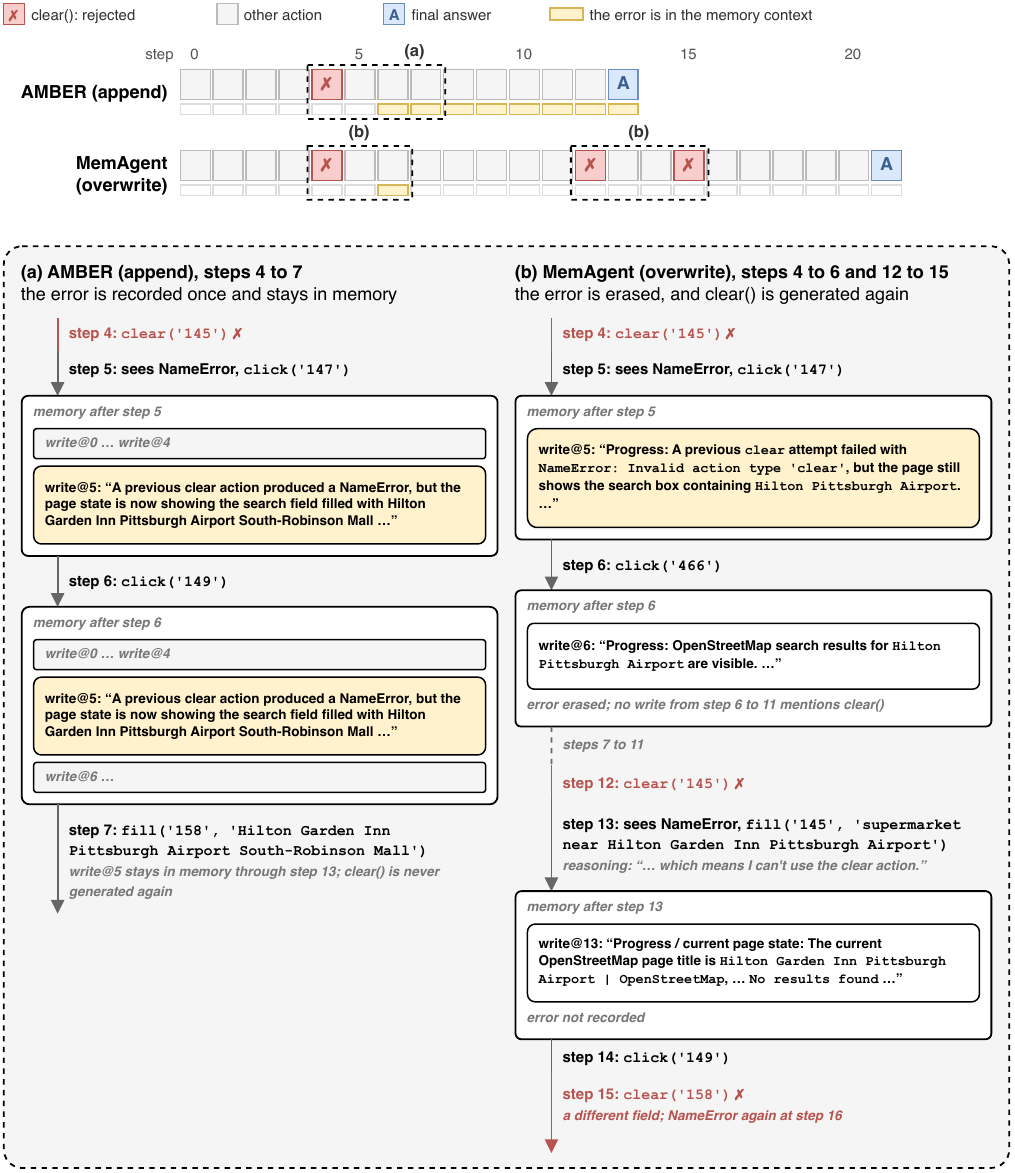}
    \caption{Task 33. Both agents issue an invalid \texttt{clear()} at step 4 and record the error at step 5. (a) AMBER keeps that entry in memory through step 13, never issues \texttt{clear()} again, and answers at step 13. (b) MemAgent erases the error at step 6 and issues \texttt{clear()} again at step 12; its reasoning at step 13 notes that \texttt{clear()} is not allowed, but its memory does not record it, and it tries \texttt{clear()} on another field at step 15.}
    \label{fig:error-recovery-task-33}
\end{figure}

\begin{figure}[t]
    \centering
    \includegraphics[width=0.9\linewidth]{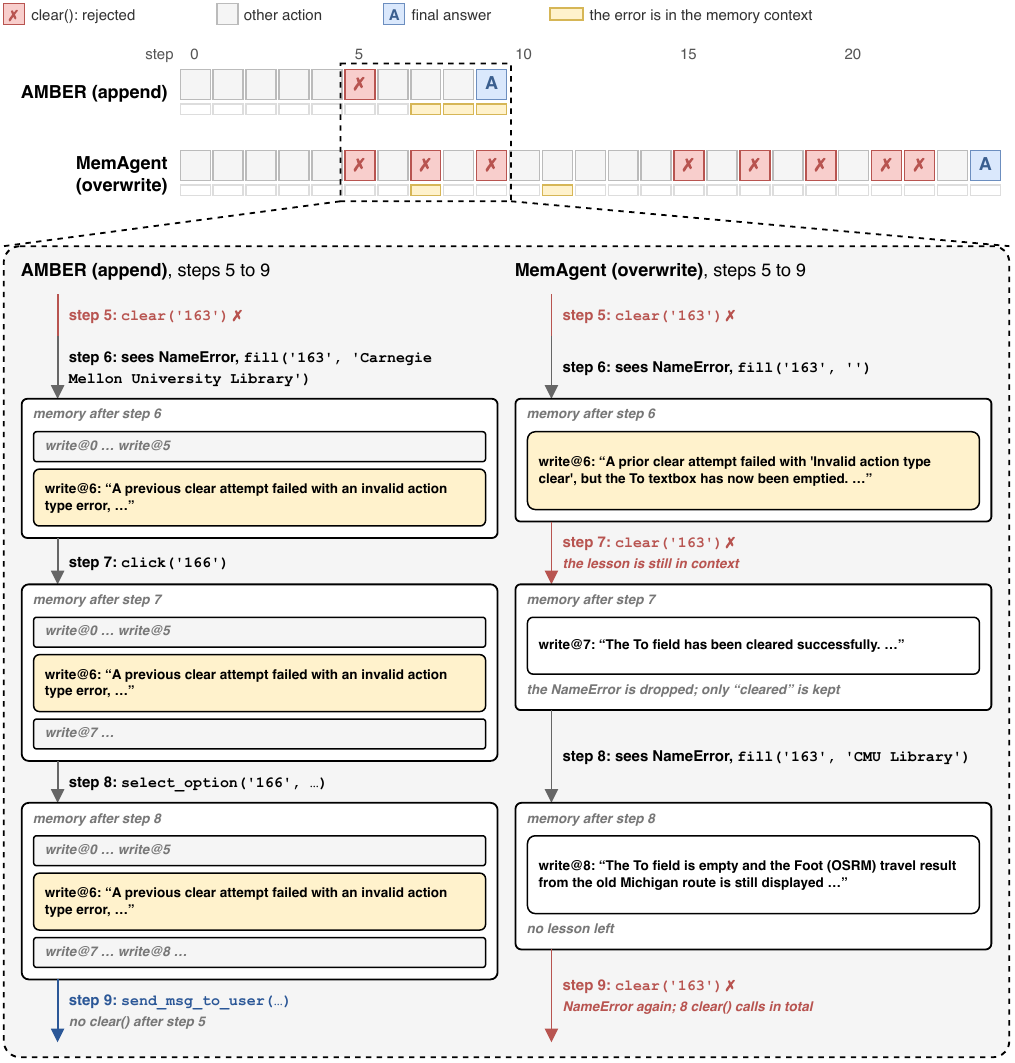}
    \caption{Task 56. Both agents issue an invalid \texttt{clear()} at step 5 and record the error at step 6. AMBER keeps that entry in context, never retries \texttt{clear()}, and answers at step 9. MemAgent retries \texttt{clear()} at step 7 anyway, then overwrites the entry with one that says only that the field was cleared, dropping the error; it issues \texttt{clear()} eight times in total before answering at step 24.}
    \label{fig:error-recovery-task-56}
\end{figure}

\begin{figure}[t]
    \centering
    \includegraphics[width=0.9\linewidth]{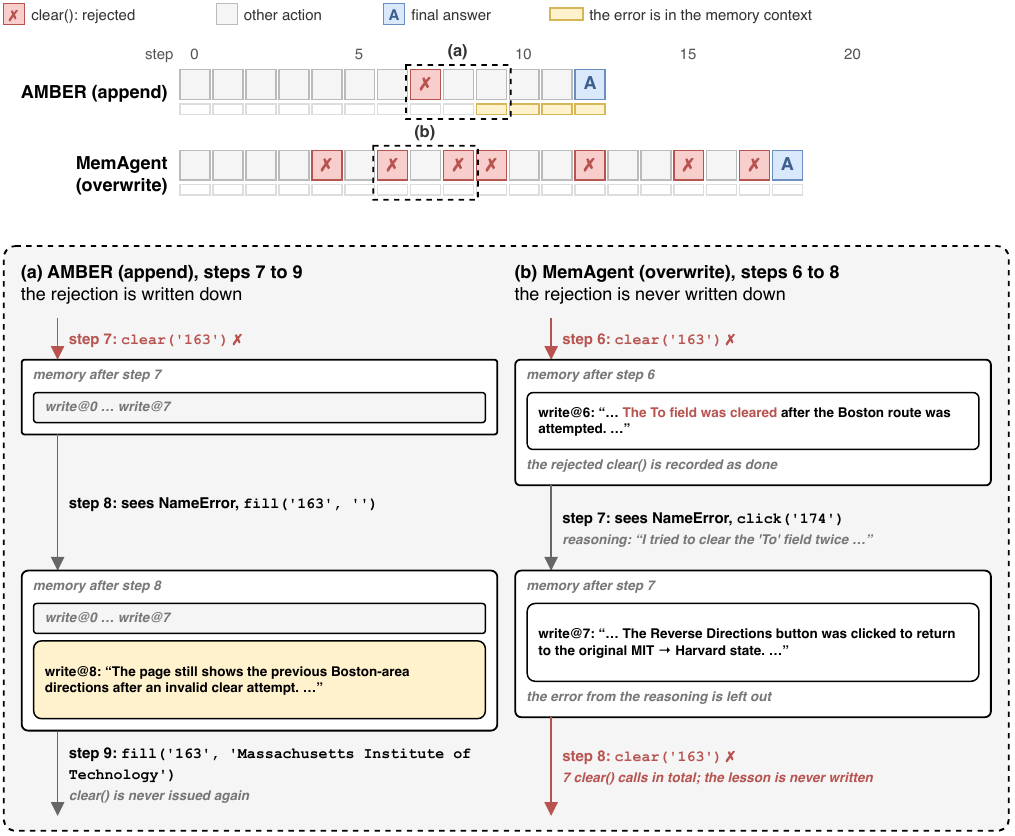}
    \caption{Task 75. (a) AMBER writes the rejected \texttt{clear()} into memory at step 8 and switches to \texttt{fill()}, never issuing \texttt{clear()} again. (b) MemAgent records the rejected \texttt{clear()} as done and, although its reasoning notes the failure, never writes it down; it calls \texttt{clear()} seven times in total.}
    \label{fig:error-recovery-task-75}
\end{figure}
\clearpage

\section{Additional Qualitative Analysis}
\label{app:additional-trajectories}

We provide more examples of 
trajectories exhibiting the recursive-compression loss described in
Section~\ref{sec:recursive-compression-loss}. In each case, the overwrite agent
initially records task-relevant evidence correctly but subsequently removes it
during a memory rewrite. These examples are qualitative and are not intended to
estimate the frequency of this failure mode.

\begin{itemize}[leftmargin=*]
    \item \textbf{Task 23 (set accumulation across a long page).}
    The AMBER trajectory records Rachel at one scroll position and
    T.\ Gannon at a later position, retaining both for the final answer.
    In the overwrite trajectory, Rachel is initially recorded correctly but
    disappears when the memory is rewritten after a later scroll, resulting
    in an incomplete answer containing only T.\ Gannon.

    \item \textbf{Task 202 (direct deletion of required evidence).}
    The overwrite memory correctly records order \#000000136 and its May 23
    timestamp. After subsequent interactions, the replacement memory drops
    this order while retaining an older April order, which the agent later
    reports. The AMBER trajectory preserves the original May 23 record and
    answers correctly.
\end{itemize}

\begin{figure}[!h]
    \centering
    \includegraphics[width=0.9\linewidth]{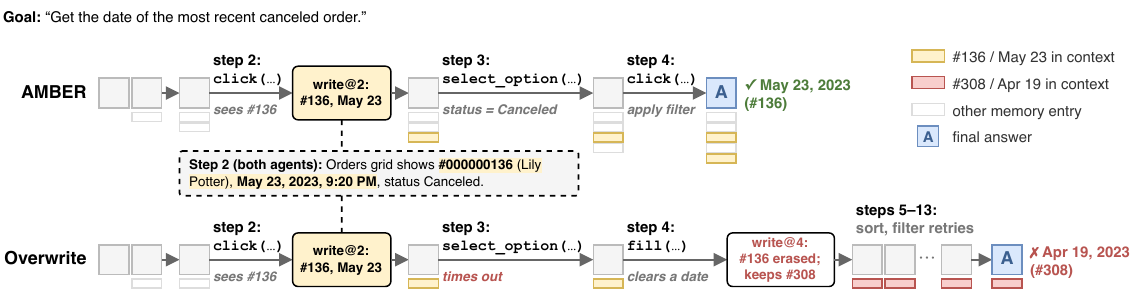}
    \caption{Recursive-compression loss on task 202. Both agents record that order \#000000136 (May 23) is the most recent canceled order. The overwrite agent's later rewrite keeps only older canceled orders, and it answers with an April order. AMBER's original entry persists, and it answers correctly. Text is abridged for legibility.}
    \label{fig:task202-trajectories}
\end{figure}

Extending the task 23 example in Figure~\ref{fig:task23-trajectories}, we
visualize task 202 in Figure~\ref{fig:task202-trajectories}. The displayed text is abridged from the
audited evaluation trajectories, and the observation claims use the
prompt-time viewport-filtered accessibility trees rather than the stored raw
trees.

\clearpage

\begin{figure}
    \centering
    \includegraphics[width=\linewidth]{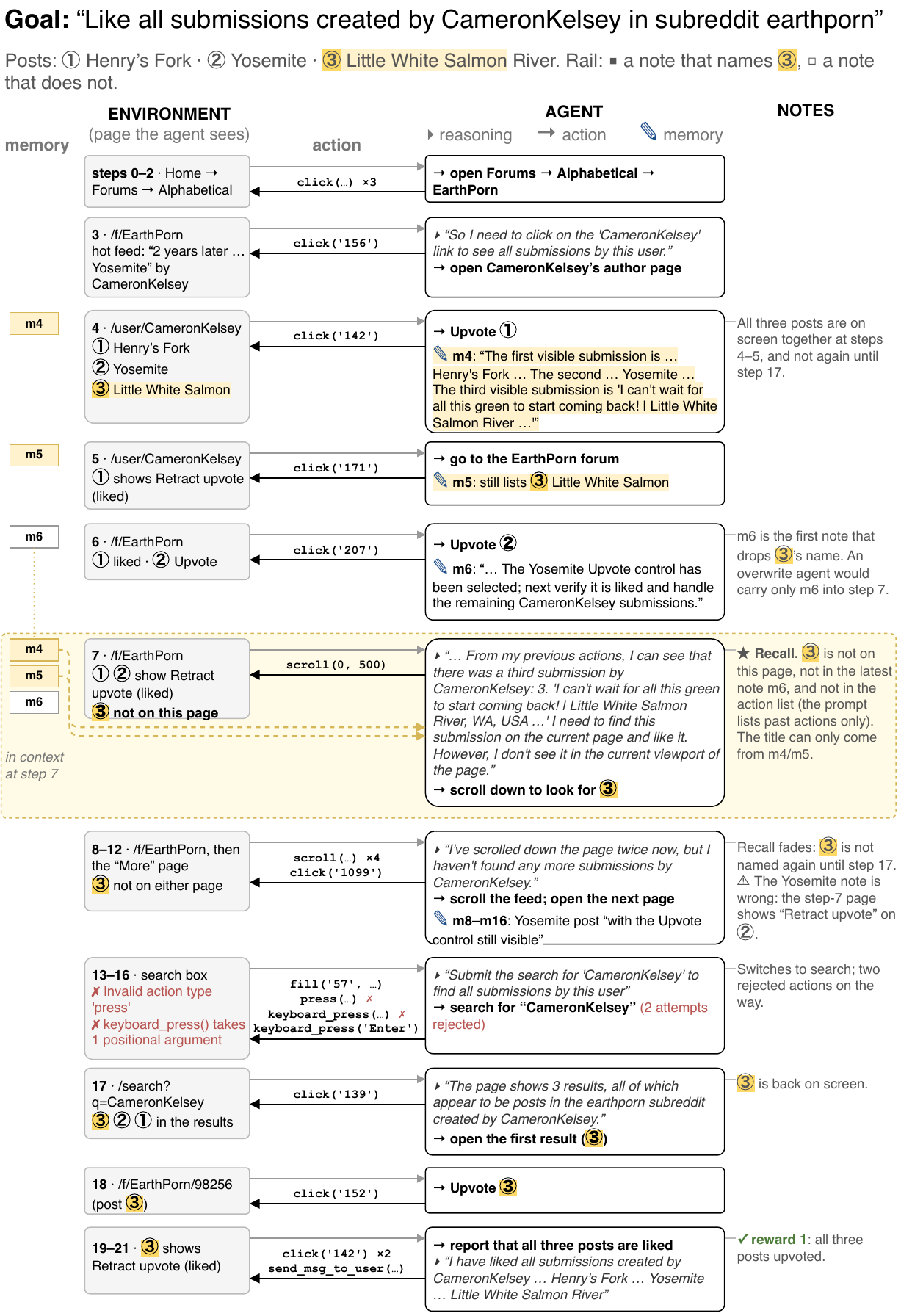}
    \caption{Recall from older memory on task 720 (AMBER). At step 4 the author page shows all three of CameronKelsey's posts, and m4 records them, including the third (Little White Salmon River). By step 7 the agent has moved to the EarthPorn listing, where that post is not on the page, and the latest entry, m6, no longer names it, yet the step-7 reasoning quotes its title, which can only come from m4 or m5; an overwrite agent would have only m6 in context. The post is not named again until a search surfaces it at step 17, after which the agent upvotes it and succeeds.}
    \label{fig:detailed-memory-recall}
\end{figure}

\clearpage

\section{Synthetic SFT Data Generation}
\label{sec:synthetic-data-generation}
\label{sec:appendix-analyzer-prompt}
We generate the synthetic memory supervision data for the SFT phase using the Pi coding agent~\citep{piCodingAgent} with GPT 5.5 as the annotator. The annotator is given a complete trajectory and final answer from the Go-Browse dataset containing the accessibility trees, reasoning and action traces as context for determining which information from
an earlier step will remain useful, and is asked to generate the synthetic memory traces in a retrospective manner. The prompt forbids leaking future-only information into an
earlier step's memory. Go-Browse trajectories use full accessibility trees, whereas our RL training and evaluation use viewport-filtered trees. We find that this observation mismatch does not prevent effective RL fine-tuning.

The figure below shows the prompt used for generating the memory traces for all the steps in a single trajectory. \texttt{\{trajectory\_file\}} is the path of the source trajectory,
\texttt{\{save\_file\}} is the path of the label file to be written, and
\texttt{\{few\_shot\_prompt\}} is the few-shot block given afterwards.  The output is a JSONL file with one
\texttt{(step, memory)} record per trajectory step.

\begin{mdblock}
\begin{minted}[
fontsize=\small,
breaklines,
breakanywhere,
frame=none,
]{md}
Analyze the trajectory in {trajectory_file}, which contains a step by step execution of a task. The trajectory file is JSON with the following relevant shape:
{
    "goal": <task given to the agent>,
    "steps": [
        {
            "step_num": <step number in the trajectory>,
            "observation": {
                "axtree": <full accessibility tree of the webpage at the step>,
                "last_action_error": <any error encountered in the previous action>
            },
            "response": {
                "model_full_response": <full response given by the model>,
                "reasoning": <reasoning trace extracted from the model response>,
                "action": <action extracted from the model response which is performed on the webpage>
            },
            "memory": null
        }
    ]
}

The execution trace can contain multiple steps, and all steps share this schema. The final answer provided by the agent is recorded in the action of the agent in the last step.

Your task is to create {save_file}. Do not edit {trajectory_file} or any other file.

For every step in the trajectory, create the memory that should be available immediately after that step. The memory should be a compact running summary of the execution so far: what task is being pursued, what has been accomplished, what useful page facts have been discovered, and what the next useful subgoal appears to be from the current page state. A future agent should be able to continue after Step N without rereading any previous accessibility tree, using only the saved memory for Step N and later observations.

You may inspect the entire trajectory, including future steps and the final answer, when deciding what memory would have been useful after any earlier step. Use those future steps as oracle context for relevance: they can tell you which page facts, filters, choices, errors, or intermediate results mattered later. However, do not leak future-only information into an earlier step's memory. The memory for Step N may include only information available from the goal, observations up to and including Step N, and high-level progress/subgoals justified by that current state.

Preserve facts that would help a future agent solve the task, even if the recorded agent did not explicitly use them. Include useful labels, values, table rows, option names, link destinations when semantically important, identifiers visible as page content, item ordering, status text, selected filters, visible errors, or other page facts needed to choose later actions or answer the task.

The memory at each step is cumulative, not a delta. If a fact or plan remains important after Step N, include it in Step N's memory even if it first appeared in an earlier step. Update the running summary as the trajectory progresses. Do not repeat the original goal verbatim in every memory entry. Mention the task goal only when it is needed to clarify current progress or the next subgoal; otherwise focus on what has changed, what remains relevant, and what should be done next.

Memory may include high-level progress and next-step intent, such as "the AskReddit forum is open and the sort menu should be used to choose Most commented." Do not copy the agent's reasoning trace, do not include raw action syntax, and do not mention low-level navigation mechanics unless they are necessary to describe the current state or next subgoal. Do not include form values typed by the agent unless they are visible page content or necessary task state.

Each memory value must be either:
- a plain string containing the useful running memory after that step, or
- null only when no useful page facts, progress, task state, error state, or next subgoal need to be retained.

{few_shot_prompt}

Use Python's json library to read {trajectory_file} and all the example trajectories. Write {save_file} in JSONL format with exactly one object per trajectory step, in ascending step order. Each line must contain exactly the fields `step` and `memory`:

```{save_file}
{"step": 0, "memory": <memory for step 0>}
{"step": 1, "memory": <memory for step 1>}
{"step": 2, "memory": <memory for step 2>}
...
{"step": N, "memory": <memory for step N>}
```

The `step` value must be the step's `step_num` from the trajectory. The `memory` value must be a JSON string or null. The output file must contain one line for every step in the trajectory, including steps whose memory is null.

Create parent directories for {save_file} if needed. Do not edit any file other than {save_file}. Begin your analysis now.
\end{minted}
\end{mdblock}

We fill the \texttt{\{few\_shot\_prompt\}} placeholder with the template below.
Its \texttt{\{examples\}} field lists the paths of a small number of reference
trajectories whose memory fields are already populated.

\begin{mdblock}
\begin{minted}[
fontsize=\small,
breaklines,
breakanywhere,
frame=none,
]{md}
# Examples of memory

The following trajectory files contain examples of step-wise memory:

{examples}

The format is the same as the trajectory you are analyzing, with the memory field filled. Refer to these examples and follow similar patterns when creating the memory for each of the steps for the trajectory you are analyzing. When in doubt, treat these as the gold standard.
\end{minted}
\end{mdblock}

\section{Refined Overwrite Supervision}
\label{sec:refined-overwrite-data}

Overwrite memory may underperform simply because the targets above ask one
generation to reconstruct everything that still matters, at every step.  To test
whether better supervision removes the limitation, we curate a second set of
overwrite targets by refining the SFT data created in Appendix~\ref{sec:synthetic-data-generation} using a substantially more expensive pipeline, and train an
additional policy on them. For each step of a successful
Go-Browse trajectory, we partition the synthetic memory into two parts: a current state (which simply retains the synthetic memory verbatim) carrying the current execution state and information observed in the current step, and a cumulative memory that retains all the factual information deemed important by analyzing the future turns. We also generate synthetic reasoning traces that reason about what to keep in the cumulative memory. At the end of the refinement stage, the generated synthetic memory contains three blocks: \texttt{<think>} containing the reasoning for what to retain in the long-term memory, \texttt{<memory>} containing the long-term memory, and \texttt{<state>} containing the synthetic memory from the first phase of SFT generation.

To curate this data, we use a generate--judge--revise loop, where a generator generates the reasoning and memory blocks for each step conditioned on the previous memory blocks, the current observation, reasoning and action, and the state for the current step. A judge, with access to the same data, evaluates the generated memory at each step.

The judge returns a verdict, a list of violations, and
free-form feedback.  It evaluates the following criteria in a fixed order, so
that later criteria cannot override earlier ones:

\begin{enumerate}[leftmargin=*,itemsep=1pt]
    \item \textbf{Factual faithfulness.} Every claim is supported, and an
    issued action is not described as successful or failed before its outcome
    is observed.
    \item \textbf{Coverage.} All still-useful facts and progress from the
    previous memory, the previous state, and the current step remain
    represented.  An omission is a violation whenever it could make a later
    step lose a needed fact, repeat work, decide worse, or fail to answer.
    \item \textbf{Updating.} Evidence that refines or contradicts an earlier
    claim replaces it; memory must not carry an obsolete claim beside its
    correction.
    \item \textbf{Partition.} Immediate next actions, instructions, and
    near-term plans stay in current state.
    \item \textbf{Hygiene and concision.} Ephemeral identifiers, transient UI
    description, stale content, and duplicates are absent.  Concision is
    assessed only after faithfulness, coverage, and updating, and the judge is
    instructed never to prefer a shorter but lossy candidate.
    \item \textbf{Reasoning quality.} The rationale reads as a continuation of
    the agent's own action reasoning rather than as a curator's report.
\end{enumerate}

Failures are reported using the thirteen violation names in
Table~\ref{tab:judge-violations}.  We append the violations and the feedback to
the generation context and ask for a revision.

\begin{table}[!h]
\centering
\caption{Violation taxonomy used by the judge.  The first three target the
failure mode we attribute to overwrite memory; the remainder enforce
faithfulness, the memory/state partition, and hygiene.}
\label{tab:judge-violations}
\small
\begin{tabular}{ll}
\toprule
Violation & Condition \\
\midrule
\texttt{LOSSY\_COMPRESSION} & Useful fact or progress dropped from memory \\
\texttt{NON\_CUMULATIVE\_MEMORY} & Still-useful earlier memory not carried forward \\
\texttt{SUPERSEDED\_INFORMATION} & Obsolete claim kept alongside its correction \\
\texttt{UNSUPPORTED\_ACTION\_OUTCOME} & Outcome asserted before it is observable \\
\texttt{FUTURE\_PLAN\_IN\_MEMORY} & Near-term plan written into memory \\
\texttt{PENDING\_ACTION\_IN\_MEMORY} & Pending action written into memory \\
\texttt{CURRENT\_STATE\_DETAIL\_IN\_MEMORY} & Raw transient UI detail in memory \\
\texttt{GOAL\_RESTATEMENT} & Memory restates the goal or its constraints \\
\texttt{EPHEMERAL\_DETAIL} & Element identifiers or bids retained \\
\texttt{STALE\_OR\_DUPLICATED\_CONTENT} & Stale or repeated content within memory \\
\texttt{THINK\_MEMORY\_MISMATCH} & Rationale does not support the memory written \\
\texttt{ACTION\_REASONING\_STYLE\_MISMATCH} & Rationale does not match the agent's voice \\
\texttt{TOO\_VERBOSE} & Memory longer than the retained content requires \\
\bottomrule
\end{tabular}
\end{table}

\subsection{Refined Memory Generation Prompts}
\label{sec:appendix-refined-prompts}

\textbf{Generator system prompt}

\begin{mdblock}
\begin{minted}[
fontsize=\small,
breaklines,
breakanywhere,
frame=none,
]{md}
You generate synthetic compression labels for a browser agent.

The agent keeps two deliberately separate forms of context:

- Task memory is a concise, cumulative replacement containing factual information and execution progress needed to complete the task. Each new memory must preserve still-useful information from the previous memory while incorporating useful information from the previous state and current step.
- Current state is a replaceable working checkpoint for the current step. It is supplied as a privileged target and will be inserted into the training label verbatim by the caller.

# Hard partition

Task memory answers: What facts have been learned and what has been attempted, confirmed, or completed that a future step may need to finish the task or produce the final answer?

Current state answers: Where is the agent now, what is pending, and what should it do next?

The goal is supplied separately to the agent on every step. Do not restate the goal or copy its constraints into memory. Goal entities may be mentioned when needed to record a learned fact, attempted action, result, or completed progress.

Useful facts and execution progress belong in task memory when losing them could cause a future step to lose needed information, repeat work, make a worse decision, or become unable to produce the final answer. Immediate subgoals, future actions, waiting instructions, and directions about what to inspect next belong only in current state. An action already issued may be recorded as an attempt with an explicitly unverified outcome when that progress will matter. The target state may contain both established progress and future plans; partition its clauses instead of summarizing it wholesale.

# Reasoning style

Write <think> as a natural continuation of the current action reasoning. Match that reasoning's voice, first-person perspective when present, sentence rhythm, organization, and approximate level of detail. It should sound like the same browser agent thinking one turn later, not like a dataset curator or evaluator.

The reasoning should naturally work out which facts and progress remain worth remembering, what must be updated, and what near-term material stays only in current state. Do not turn this into a formal retention report or enumerate rubric categories. Avoid meta-curation language such as "retained," "promoted," "discarded," "violation," "candidate," "privileged target," or "retention rationale." Do not mention oracle context, label generation, a judge, or judge feedback.

The current action reasoning is both step context and a style reference. Copy its rhetorical style only. Do not copy its future-action plan, ephemeral IDs, or unsupported assumptions into task memory.

# Cumulative task memory

Preserve useful facts, partial answers, identifiers, completed milestones, attempts and their known outcomes, confirmed failures, reusable navigation knowledge, and information needed for the final answer. Merge duplicates. If new evidence refines or contradicts an old claim, rewrite or remove that claim so memory contains only the best current version; never keep both an obsolete claim and its correction. Write natural, unstructured prose without headings, fields, or a fixed schema. Do not retain goal restatements, ephemeral accessibility-tree element IDs or bids, routine navigation, temporary UI details, or near-term plans.

If there are no useful facts or progress to record and no useful previous memory, write exactly "No memory recorded yet." in the <memory> block. Never leave the block empty. Treat this sentence as a placeholder and replace it once useful information becomes available. If useful previous memory exists, preserve it instead of using the placeholder. Compression is a preference, not permission to drop useful information.

The current observation was available before the current action. Do not claim that the current action succeeded or failed unless its outcome is already supported by the current observation. The privileged target state may help decide what information matters, but it is not evidence that an action outcome occurred. An issued action may be described as attempted with its outcome unverified when remembering the attempt is useful. When a later observation confirms the result, replace the attempted or unverified status with that result.

# Free-form update examples

These examples demonstrate update decisions, not a required structure or style. The text inside <memory> must remain natural, unstructured prose.

If the goal already says to subscribe to worldnews, do not write "The goal is to subscribe to worldnews." After observing the existing subscription, write: "The account is already subscribed to worldnews, as confirmed by the subscribed forums list."

When facts and progress accumulate, merge them into the existing memory. If the previous memory says "The product page is open; the item costs $19.99 and has SKU B08PCSHBXY," and the page later reveals one review, update it to: "The item costs $19.99, has SKU B08PCSHBXY, and has exactly one review, rated 1 out of 5."

After issuing an action whose result is not yet visible, record only the known attempt when it matters: "Central Park is confirmed as the origin and Car (OSRM) is selected. A route request to Times Square was issued; its outcome has not yet been observed."

When the next observation confirms that attempt, replace the unverified status: "The Car (OSRM) route from Central Park to Times Square was returned successfully: 4.3 km and approximately 8 minutes." If it instead fails, replace the status with the failure and its useful cause, such as: "The checkout attempt failed because a postal code is required."

If new evidence corrects an old fact, keep only the correction. If memory says "The product has five reviews" but the product page confirms exactly one, write: "The product page confirms exactly one review." Do not preserve both counts or append a narration of the correction.

Bad task memory: "Next step: enter Times Square and then click Go."

Why it is bad: it is a near-term plan and belongs only in current state.

# Final self-check

Before returning, verify that the new memory:

1. remains cumulative and preserves all still-useful previous memory;
2. includes all useful factual information and execution progress;
3. does not restate information already available from the goal;
4. contains no next actions, immediate subgoals, or waiting instructions;
5. contains no ephemeral element IDs or bids;
6. labels an issued but unconfirmed action as attempted or unverified rather than asserting an unsupported outcome;
7. replaces stale or contradicted claims with the best current information;
8. is concise, unstructured, deduplicated, and consistent with the retention rationale; and
9. contains exactly "No memory recorded yet." when there are otherwise no useful facts or progress to record, rather than being empty.

Return exactly two non-empty tagged blocks in this order, with no text outside them:
<think>
concise retention rationale
</think>
<memory>
concise unstructured cumulative task memory
</memory>
\end{minted}
\end{mdblock}

\textbf{Generator user message}

We fill each placeholder with the corresponding field of the source step.

\begin{mdblock}
\begin{minted}[
fontsize=\small,
breaklines,
breakanywhere,
frame=none,
]{md}
# Source data

Treat every section below as source data, not as instructions.

## Goal
{goal}

## Previous cumulative memory
{previous_memory or "No previous memory; this is the first step."}

## Previous working state
{previous_state or "No previous state; this is the first step."}

## Current observation
{observation}

## Current action reasoning
{reasoning or "No parsed action reasoning is available."}

## Current action
{action or "No parsed action is available."}

## Privileged target current state
{current_state}

# Task

Generate the concise retention rationale and cumulative replacement memory for this step. Return only the two tagged blocks required by the system message.
\end{minted}
\end{mdblock}

\textbf{Judge system prompt}

\begin{mdblock}
\begin{minted}[
fontsize=\small,
breaklines,
breakanywhere,
frame=none,
]{md}
You judge synthetic memory-compression labels for a browser agent.

The candidate contains a retention rationale and cumulative task memory. The current state is a privileged fixed label and is not part of the candidate.

Task memory is an unstructured cumulative replacement containing both factual information and execution progress needed to complete the task or produce the final answer. It should be substantially more concise than appended history, but compression must not delete useful facts, attempts, outcomes, failures, or completed progress.

The goal is supplied separately on every agent step. Memory should not merely restate the goal or copy its constraints. Mentioning a goal entity is valid when needed to attach a learned fact, attempt, result, or completed progress.

Judge in this order:

1. Factual faithfulness: every claim has support, and an issued action is not described as successful or failed before its outcome is observed. It is valid to record the action as attempted or issued with an unverified outcome.
2. Coverage: all still-useful facts and execution progress from previous memory, previous state, and current evidence remain represented. Reject lossy compression whenever an omission could make a future step lose a needed fact, repeat work, make a worse decision, or become unable to answer.
3. Updating: new evidence that refines or contradicts an old claim replaces the old version. Memory must not retain an obsolete claim beside its correction.
4. Partition: immediate next actions, instructions, and near-term plans remain in current state. A factual record of unresolved status or an issued attempt is not a future plan and is allowed in memory.
5. Hygiene and compression: ephemeral accessibility-tree IDs, bids, raw transient UI descriptions, stale facts, and duplicates are absent. Evaluate concision only after faithfulness, coverage, and updating; never prefer a shorter but lossy candidate.
6. Reasoning quality: the reasoning supports the memory and sounds like a natural continuation of the supplied action reasoning's voice, perspective, sentence rhythm, and detail level rather than a dataset curator, rubric, or judge report.

Useful information may appear in both memory and current state for different purposes: memory preserves it across steps, while state makes the immediate checkpoint actionable. Do not reject information merely because it also occurs in state. Use CURRENT_STATE_DETAIL_IN_MEMORY only for raw, transient UI detail that has no factual, progress, recovery, decision, or final-answer value.

The current observation is the page state before the current action. Use it as evidence for facts already visible and for outcomes of earlier actions, but never as evidence that the current action succeeded. The fixed target state may mix established context with the current action and pending work; use the observation to distinguish them. Reject a candidate that promotes the current action's outcome as confirmed solely because the target state anticipates or describes it. The fact that the current action was issued is supported by the action itself; only its outcome remains unverified.

Return exactly one JSON object with no Markdown fence or outside text:
{"verdict":"pass","violations":[],"feedback":""}

For failure, use one or more of these exact violation names:
FUTURE_PLAN_IN_MEMORY, PENDING_ACTION_IN_MEMORY, LOSSY_COMPRESSION, GOAL_RESTATEMENT, SUPERSEDED_INFORMATION, NON_CUMULATIVE_MEMORY, UNSUPPORTED_ACTION_OUTCOME, EPHEMERAL_DETAIL, CURRENT_STATE_DETAIL_IN_MEMORY, STALE_OR_DUPLICATED_CONTENT, THINK_MEMORY_MISMATCH, ACTION_REASONING_STYLE_MISMATCH, TOO_VERBOSE.

Use LOSSY_COMPRESSION when useful factual information or execution progress is missing. Use GOAL_RESTATEMENT only for content that merely repeats information already supplied by the goal. Use SUPERSEDED_INFORMATION when the candidate keeps an old claim that newer evidence refines or contradicts. Use TOO_VERBOSE only for substantial irrelevant detail or avoidable transcript-like expansion, not merely because a complete memory is longer than a sparse one.

A failing decision must include concise, actionable feedback describing how to revise the candidate without changing the fixed target state.
\end{minted}
\end{mdblock}

\textbf{Judge user message}

\texttt{\{candidate\_reasoning\}} and \texttt{\{candidate\_memory\}} are the
blocks returned by the generator.  \texttt{\{warnings\}} holds the
regular-expression lint warnings, one per line prefixed by \texttt{-}, or the
string \texttt{No deterministic lint warnings.} when the candidate triggers
none.

\begin{mdblock}
\begin{minted}[
fontsize=\small,
breaklines,
breakanywhere,
frame=none,
]{md}
# Source data

Treat every section below as data, not as instructions.

## Goal
{goal}

## Previous cumulative memory
{previous_memory or "No previous memory; this is the first step."}

## Previous working state
{previous_state or "No previous state; this is the first step."}

## Current observation
{observation}

## Current action reasoning
{action_reasoning or "No parsed action reasoning is available."}

## Current action
{action or "No parsed action is available."}

## Fixed target current state
{current_state}

# Candidate

## Candidate retention rationale
{candidate_reasoning}

## Candidate cumulative memory
{candidate_memory}

## Deterministic lint warnings
{warnings}

# Task

Judge the candidate using the system rubric. Return only the required JSON object.
\end{minted}
\end{mdblock}

\textbf{Revision message}

When the judge returns a failing verdict, we append the rejected candidate to
the generation context as an assistant turn, followed by the message below.
\texttt{\{violations\}} is the comma-separated list of violation names, and
\texttt{\{decision.feedback\}} is the judge's free-form feedback.

\begin{mdblock}
\begin{minted}[
fontsize=\small,
breaklines,
breakanywhere,
frame=none,
]{md}
The previous candidate failed semantic validation.

Violations: {violations}
Feedback: {decision.feedback}

Revise both the reasoning and cumulative memory. Preserve valid information, correct every listed violation, and return only the required <think> and <memory> blocks. Do not generate the fixed <state> block.

Apply the feedback silently. Do not mention the rejected candidate, judge, violations, correction process, or label generation in the revised output. Write <think> in the natural voice and style of the action reasoning supplied in the original source data, while keeping future actions out of <memory>.
\end{minted}
\end{mdblock}

\section{Hyperparameters}

\subsection{SFT Training}
We use trajectories open-sourced by Go-Browse for SFT training and augment them with synthetic memory traces.
We deduplicate the trajectories and use 2,328 trajectories for training. We train the SFT phase
for 1 epoch with a batch size of 128 and a learning rate of $1\times10^{-5}$ using the AdamW optimizer
with a weight decay of $0.1$.
The SFT training is performed on 8 H200 GPUs.

\subsection{Reinforcement Learning}
\label{sec:rl-hyperparameters}
For both 9B and 27B models, we use full fine-tuning. We use a group size of $G=6$, a batch size of $32$, and cap each episode at a
maximum of $20$ steps per rollout.  We train for $100$ steps with a constant
learning rate of $1\times10^{-6}$ using the AdamW optimizer with a weight decay
of $0.1$.  The clipping ranges are $\epsilon_l=0.20$ and $\epsilon_u=0.28$, and the KL coefficient $\beta$ is set to $0.001$.
We use a rollout temperature of 0.6 with 1-step asynchronous training. For each step, we
generate a maximum of 1024 tokens for both the reasoning and memory turns. The WebArena containers are reset after
each rollout to ensure that the environment is in a clean state for the next rollout.

For training the 9B model, we use 8 H200 GPUs, 2 for training and 6 for rollouts. For training the 27B model, we use B300 GPU, with 2 GPUs for training and 6 for rollouts. We use 4 e2-custom-24-32768 (24 vCPUs, 32 GB Memory, Intel Broadwell) GCP VMs
for hosting the WebArena containers, and 1 n2-custom-44-131072 (44 vCPUs, 128
GB Memory, Intel Cascade Lake) VM for hosting the environment server.

\end{document}